\documentclass[lettersize,journal,twoside]{IEEEtran}
\usepackage{amsmath,amsfonts,amssymb}
\usepackage{algorithmic}
\usepackage{algorithm}
\usepackage{array}
\usepackage[caption=false,font=normalsize,labelfont=sf,textfont=sf]{subfig}
\usepackage{textcomp}
\usepackage{stfloats}
\usepackage{url}
\usepackage{verbatim}
\usepackage{graphicx}
\usepackage{cite}
\usepackage[export]{adjustbox}
\usepackage[table,xcdraw]{xcolor}
\usepackage{color}
\usepackage[hidelinks]{hyperref}
\definecolor{linkcolor}{HTML}{ec008c}
\hypersetup{
	colorlinks=true,
	linkcolor=red,
	citecolor=red,
}
\usepackage{multirow}
\usepackage{booktabs}
\usepackage{orcidlink}
\usepackage{threeparttable}
\usepackage{pifont}

\newcommand\liehat[1]{\left[ #1 \right]_\times}

\newcommand\mlcomment[1]{\iffalse #1 \fi}
\newcommand\bsm[1]{\boldsymbol{\mathrm{#1}}}
\newcommand\transform[2]{{\bsm{T}_{#1}^{#2}}}

\newcommand\rotation[2]{{\bsm{R}_{#1}^{#2}}}
\newcommand\rotationhat[2]{{\hat{\bsm{R}}_{#1}^{#2}}}

\newcommand\timeoffset[2]{{\tau_{#1}^{#2}}}
\newcommand\timeoffsethat[2]{{\hat{\tau}_{#1}^{#2}}}

\newcommand\angvel[2]{{\bsm{\omega}_{#1}^{#2}}}

\newcommand\translation[2]{{\bsm{p}_{#1}^{#2}}}
\newcommand\translationhat[2]{{\hat{\bsm{p}}_{#1}^{#2}}}

\newcommand\linvel[2]{{\bsm{v}_{#1}^{#2}}}
\newcommand\linvelhat[2]{{\hat{\bsm{v}}_{#1}^{#2}}}
\newcommand\linveltilde[2]{{\tilde{\bsm{v}}_{#1}^{#2}}}
\newcommand\linacce[2]{{\bsm{a}_{#1}^{#2}}}

\newcommand\gravity[1]{{\bsm{g}^{#1}}}
\newcommand\gravityhat[1]{{\hat{\bsm{g}}^{#1}}}

\newcommand\smallminus{{\text{-}}}
\newcommand\smallplus{{\text{+}}}
\newcommand\coordframe[1]{\underrightarrow{\mathcal{F}}_{#1}}

\usepackage{siunitx}
\begin{document}

\title{iKalibr-RGBD: Partially-Specialized Target-Free Visual-Inertial Spatiotemporal Calibration For RGBDs via Continuous-Time Velocity Estimation}

\author{
Shuolong Chen \hspace{-1mm}$^{\orcidlink{0000-0002-5283-9057}}$, Xingxing Li \hspace{-1mm}$^{\orcidlink{0000-0002-6351-9702}}$, 
Shengyu Li \hspace{-1mm}$^{\orcidlink{0000-0003-4014-2524}}$, and
Yuxuan Zhou \hspace{-1mm}$^{\orcidlink{0000-0002-5261-0009}}$
\thanks{
Manuscript received: xx xx, xxxx; Revised: xx xx, xxxx; Accepted: xx xx, xxxx. 
This work was supported by the the National Science Fund for Distinguished Young Scholars of China under Grant 42425401, the National Natural Science Foundation of China under Grant 423B240.
Corresponding author: Xingxing Li (\emph{xxli@sgg.whu.edu.cn}). 
The authors are with the School of Geodesy and Geomatics, Wuhan University, Wuhan 430070, China.}
\thanks{
The specific contributions of the authors to this work are listed in Section \emph{CRediT Authorship Contribution Statement} at the end of the article.
}
}

\markboth{Journal of \LaTeX\ Class Files,~Vol.~14, No.~8, August~2021}
{Chen \MakeLowercase{\textit{et al.}}: iKalibr-RGBD: Partially-Specialized Target-Free Visual-Inertial Spatiotemporal Calibration For RGBDs}


\maketitle

\begin{abstract}
Visual-inertial systems have been widely studied and applied in the last two decades (from the early 2000s to the present), mainly due to their low cost and power consumption, small footprint, and high availability.
Such a trend simultaneously leads to a large amount of visual-inertial calibration methods being presented, as accurate spatiotemporal parameters between sensors are essential for visual-inertial fusion.
In our previous work, i.e., \emph{iKalibr}, a continuous-time-based visual-inertial calibration method was proposed as a part of one-shot multi-sensor resilient spatiotemporal calibration.
While requiring no artificial calibration target brings considerable convenience, computationally expensive pose estimation is demanded in initialization and batch optimization, limiting its availability.
Fortunately, this can be vastly improved for the RGBDs with additional depth information, by employing mapping-free ego-velocity estimation instead of mapping-based pose estimation.
In this paper, we present an ego-velocity-estimation-based RGBD-inertial spatiotemporal calibrator, termed as \emph{iKalibr-RGBD}, which is also targetless and continuous-time-based, but computationally efficient.
The general pipeline of \emph{iKalibr-RGBD} is inherited from \emph{iKalibr}, composed of a rigorous initialization procedure and several continuous-time batch optimizations.
The implementation of \emph{iKalibr-RGBD} is open-sourced at (\url{https://github.com/Unsigned-Long/iKalibr}) to benefit the research community.
\end{abstract}

\section{Introduction}
\IEEEPARstart{V}{isual}-inertial systems, generally composed of vision sensors and inertial measurement units (IMUs), have been applied in a wide range of robotic tasks, mainly for ego-motion estimation \cite{qin2018vins,campos2021orb} and navigation \cite{carlone2018attention,huang2019visual}.
With constantly emerging new vision sensors, such as depth-aware RGBD and change-aware event cameras, visual-inertial applications become more extensive and ubiquitous.
For such visual-inertial sensor suites, accurate spatiotemporal calibration is crucial for reliable data fusion.

Visual-inertial spatiotemporal calibration, regardless of the type of vision sensor integrated in the suite, could usually be achieved by employing existing well-studied calibration methods \cite{furgale2013unified,qin2018online,fu2021high,nikolic2016non} orienting to the conventional optical cameras (capturing color or gray images), since conventional optical images can be generally provided by these mechanism-disparate vision sensors (mainly benefiting from supports of manufacturers).
Such a calibration route is also followed by our previous work \emph{iKalibr} \cite{chen2024ikalibr}, whose visual-inertial spatiotemporal calibration exactly focuses on optical vision sensors.
As a targetless, continuous-time, and open-source calibrator, \emph{iKalibr} provides significant benefits to spatiotemporal calibration.
However, significant computation cost exists in \emph{iKalibr}, especially in visual-inertial calibration, as the computationally expensive mapping-based structure from motion (SfM) \cite{schoenberger2016sfm} is required for each optical camera in calibration.
Fortunately, this can be avoided for RGBD cameras with available depth information, through partially specialized visual modeling, which is exactly the RGBD-inertial spatiotemporal calibration this work presents.
\label{key}
\begin{figure}[t]
	\centering
	\includegraphics[width=0.98\linewidth]{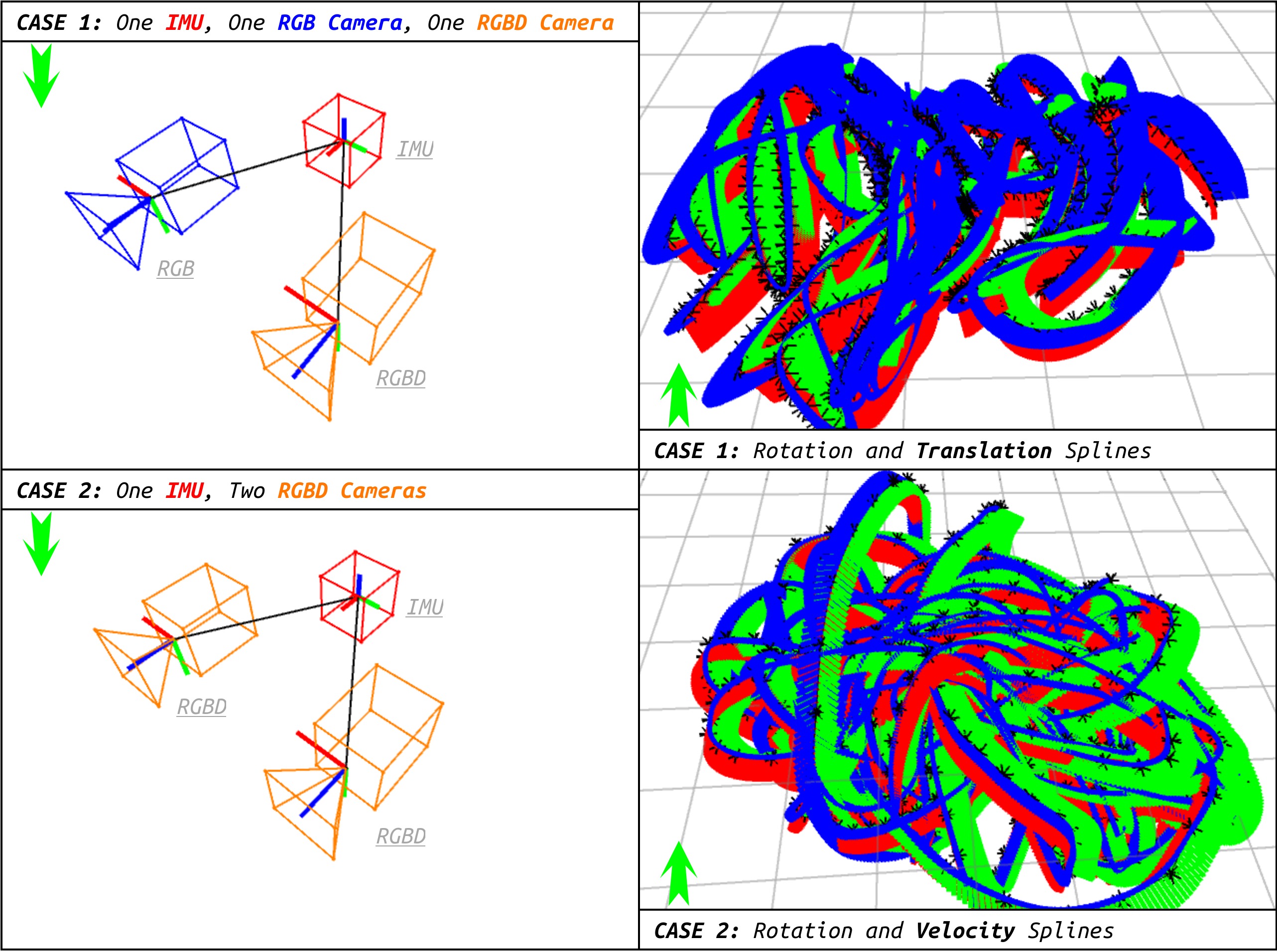}
	\caption{Runtime visualization of visual-inertial spatiotemporal calibration in \emph{iKalibr}, where two cameras and an IMU are involved. When depth information is available for all cameras (e.g., RGBDs), the velocity B-spline, instead of the translation B-spline, would be maintained in the estimator for continuous-time ego-velocity estimation.}
	\label{fig:runtime}
\end{figure}

Specifically, the commonly employed visual reprojection residual model is utilized in our previous work \cite{chen2024ikalibr} for visual-inertial calibration, whose construction relies on a trajectory (composed of time-varying rotation and translation).
This inevitably introduces mapping-based pose estimation in calibration, which needs significant computational cost in initialization and batch optimization to ensure accuracy and consistency for the visual map and trajectory.
A feasible idea to address the issue is to remodel visual residual, and this indeed works for RGBD cameras.
When visual depth information is available, mapping-free vision-only ego-velocity estimation \cite{kim20146} (composed of time-varying rotation and velocity) can be efficiently performed, capable of simultaneous spatiotemporal calibration by fusing inertial measurements.
An intuitive understanding of visual ego-velocity estimation is that the camera velocity can be implicitly inferred based on optical flow in the image plane if depth is known, see Section \ref{sect:visual_kinematic_imu_model}.
Such visual modeling is compatible with raw inertial measurements and vision-derived kinematics \cite{lu2023event} from optical flow, and can significantly reduce the computation consumption at the same time.

The proposed RGBD-inertial spatiotemporal calibration in this paper (see Fig. \ref{fig:runtime}), as an extension of \emph{iKalibr}, follows the main features of \emph{iKalibr}, namely is targetless and continuous-time-based, and innately blends into \emph{iKalibr} for one-shot resilient spatiotemporal calibration with other previously supported sensors, such as 3D mm-Wave radar, LiDAR, optical cameras, and IMUs.
The main pipeline of \emph{iKalibr-RGBD} is inherited from \emph{iKalibr}.
Specifically, the rotation B-spline is first recovered using raw angular velocities from the gyroscope.
Following that, the optical flow would be conducted to track features for extrinsic rotation initialization and camera-frame linear velocity recovery.
The recovered velocities would be ($i$) aligned with specific force from the accelerometer to initialize extrinsic translation and world-frame gravity, and ($ii$) subsequently utilized to initialize the linear velocity B-spline.
Finally, several continuous-time batch optimizations would be performed to refine all states to optimal ones by fusing sensor measurements.
\emph{iKalibr-RGBD} makes the following (potential) contributions:
\begin{enumerate}
\item We present a target-free RGBD-IMU calibration method based on continuous-time ego-velocity estimation, which is \textbf{mapping-free}, requires \textbf{low} computation consumption, and innately blends into its predecessor \emph{iKalibr} for one-shot resilient spatiotemporal calibration.

\item A rigorous dynamic initialization procedure is designed to effectively and efficiently recover \textbf{all} parameters in the estimator based on raw visual and inertial measurements.

\item Sufficient real-world experiments are performed to evaluate the proposed calibration method.
Code implementations of \emph{iKalibr-RGBD} are open-sourced, to benefit the robotic
community if possible.
\end{enumerate}

To ensure readability, only the single-RGBD single-IMU spatiotemporal calibration pipeline is presented and evaluated in this paper, as it's the most critical one to support RGBD cameras available in the one-shot multi-sensor resilient spatiotemporal calibration of \emph{iKalibr}.
Details about multi-sensor resilient calibration support in \emph{iKalibr} can be found in \cite{chen2024ikalibr}.

\section{Related Works}
This section reviews existing representative visual-inertial calibration works, as well as vision-related ego-velocity estimation involved in the proposed \emph{iKalibr-RGBD}.

\subsection{Visual-Inertial Calibration}
Artificial visual targets, such as chessboards and April Tags, are commonly utilized in the early target-based visual-inertial calibration works for efficient and accurate data association.
Based on the Kalman filter, Mirzaei et al. \cite{mirzaei2008kalman} presented a checker-board-based visual-inertial extrinsic calibration method.
Similarly relying on chessboards but employing continuous-time-based estimation, Furgale et al. \cite{furgale2013unified} proposed the well-known \emph{Kalibr} for both spatial (extrinsic) and temporal calibration of a visual-inertial sensor suite, which is the first continuous-time-based spatiotemporal calibration.
This work was subsequently extended by Huai et al. \cite{huai2022continuous} to support rolling shutter cameras besides global shutter ones.
In addition to the above target-based calibration, target-free methods are also studied, most of which are online ones integrated into real-time visual-inertial estimators.
Focusing on online temporal determination, Li et al. \cite{li2014online} presented an extended Kalman filter (EKF)-based visual-inertial odometry, where the time offset to be calibrated is treated as a part of the state vector.
Besides temporal calibration, online extrinsic determination is also studied in \cite{hartzer2022online} by Hartzer et al., which is also EKF-based but supports multiple cameras.
More recently, Huai et al. \cite{huai2022observability} presented an EKF-based visual-inertial odometry with online extrinsic, intrinsic, and temporal calibration.
In our previous work \cite{chen2024ikalibr}, we proposed a continuous-time-based and targetless spatiotemporal calibration method for integrated inertial systems, one of whose modules is visual-inertial calibration. 
Note that the above-discussed calibration methods orient to normal optical cameras, but can also apply to RGBD cameras, as RGB or gray images can be generally provided by RGBDs.

\subsection{Vision-Related Ego-Velocity Estimation}
While ego-motion estimation is essential in most robotic tasks for autonomous exploration, ego-velocity estimation is also demanded, especially in safety-critical control problems, e.g., obstacle avoidance \cite{chen2024river}.
In most filter-based ego-motion estimators \cite{geneva2020openvins,huai2022observability}, the velocity of the body is treated as a part of the state vector, and estimated together with position and attitude.
However, such mapping-based ego-motion estimators are prone to fail in highly dynamic motions or feature-poor environments.
Considering this, some recent works decouple the ego-velocity estimation from ego-motion estimation for continuous and reliable velocity acquisition.
Utilizing optical flow tracking, Kim et al. \cite{kim20146} presented a RGBD-only Kalman-filter-based ego-velocity estimator to obtain smooth velocity information.
Acquiring depth of features by substituting stereo cameras for RGBD, McGuire et al. \cite{mcguire2017efficient} proposed an efficient edge-based optical flow for velocity estimation for micro aerial vehicles with limited computing resources.
Similarly based on optical flow but powered by bio-inspired event cameras, a stereo-event-inertial ego-velocity estimator is proposed by Lu et al. \cite{lu2023event}, where depth information is obtained by event-based stereo matching and triangulation.
The event camera was also utilized by Xu et al. \cite{xu2023tight} for event-inertial ego-velocity estimation, where line-based features are fused in the estimator.
Compared with normal optical cameras, novel event cameras are more popular in velocity estimation, due to they suffer free from motion blur and have microsecond-resolution response to brightness changes.

\section{Preliminaries}
This section details the notations, sensor models, and the B-spline-based continuous-time time-varying state representation used in this paper.
\subsection{Notations}
The notation convention in this paper follows the one in our previous work \cite{chen2024ikalibr}.
Given a RGBD-inertial sensor suite, we use $\coordframe{c}$ and $\coordframe{b}$ to represent the coordinate frames of the RGBD camera and IMU, respectively.
The six-degrees-of-freedom (DoF) rigid transform from $\coordframe{c}$ to $\coordframe{b}$ is expressed as:
\begin{equation}
\small
\transform{c}{b}\triangleq
\begin{bmatrix}
\rotation{c}{b}&\translation{c}{b}\\
\bsm{0}_{1\times 3}&1
\end{bmatrix}\in\mathrm{SE(3)},\;
\mathrm{s.t.}\;
\rotation{c}{b}\in\mathrm{SO(3)}\;\mathrm{and}\;
\translation{c}{b}\in\mathbb{R}^3
\end{equation}
where $\rotation{c}{b}$ and $\translation{c}{b}$ are the rotation matrix and translation vector, respectively.
$\angvel{b}{w}$, $\linvel{b}{w}$, and $\linacce{b}{w}$ denote the angular velocity, linear velocity, and linear acceleration of $\coordframe{b}$ with respect to and parameterized in the world frame $\coordframe{w}$ ($\coordframe{w}\triangleq\coordframe{b_0}$, i.e., the first coordinate frame of $\coordframe{b}$).
The gravity vector $\gravity{w}$ with constant norm is also parameterized in $\coordframe{w}$.
Quantities with $\hat{(\cdot)}$ and $\tilde{(\cdot)}$ are estimates and noisy measurements, respectively.

\subsection{Visual Optical Flow and Inertial Model}
\label{sect:visual_kinematic_imu_model}

\begin{figure}[t]
	\centering
	\includegraphics[width=0.98\linewidth]{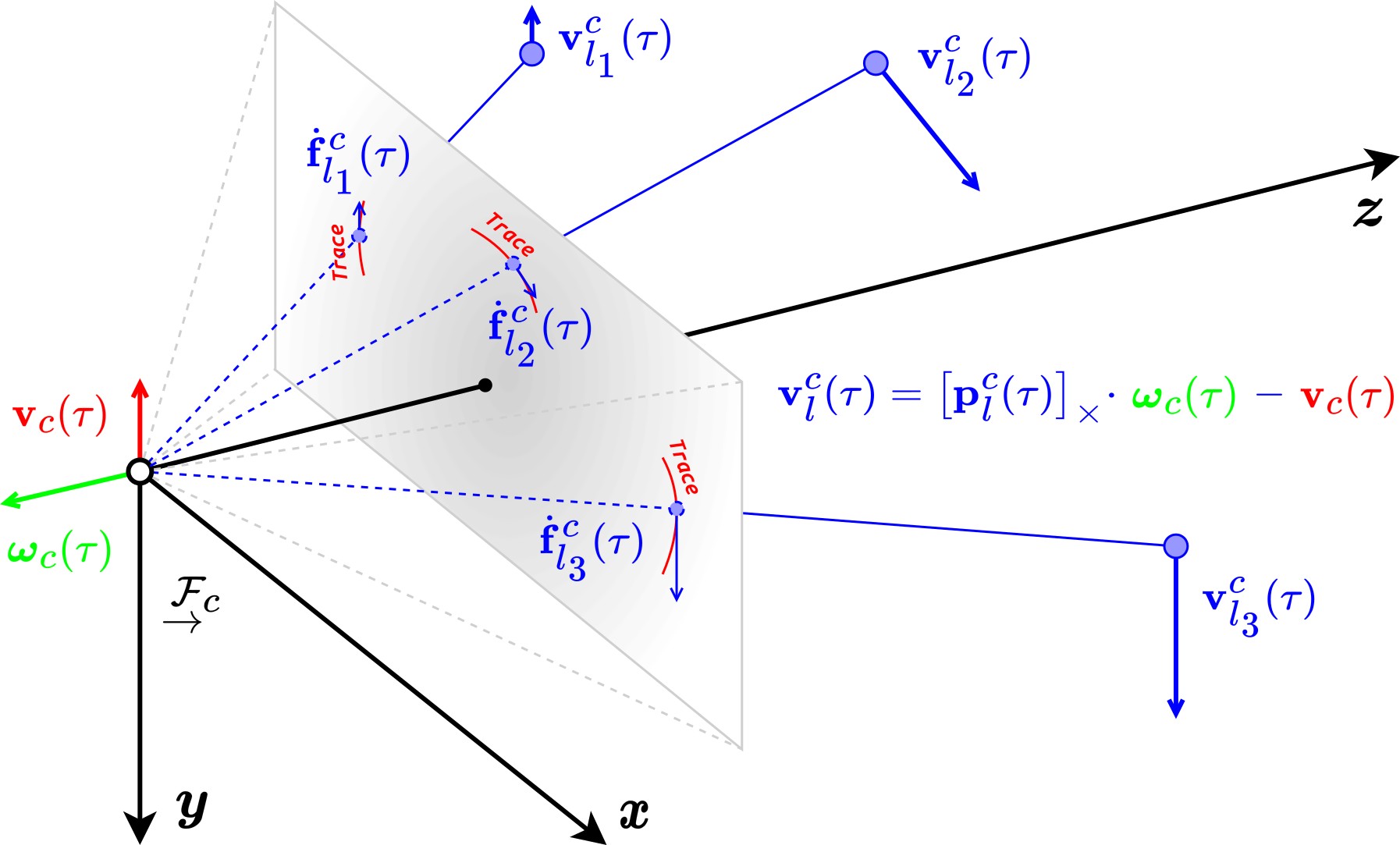}
	\caption{Illustration of first-order visual kinematics. Three-dimensional linear velocities of landmarks with respect to the camera, i.e., $\linvel{l}{c}(\tau)$, are projected onto the image plane as corresponding two-dimensional pixel velocities $\dot{{\bsm{f}}}_l^c(\tau)$.}
	\label{fig:visual_kinematics}
\end{figure}

The optical flow, defined as the projection of velocities of landmarks with respect to the camera onto the image plane, can implicitly characterize existing relative motion between the camera and the visual scenario \cite{beauchemin1995computation}.
Assume that a stationary visual landmark $\translation{l}{w}$ in $\coordframe{w}$ is continuously captured by a moving camera onto the image plane as feature $\bsm{f}_l^c(\tau)$, the corresponding first-order optical flow equation at time $\tau$ could be described as follows:
\begin{equation}
\label{equ:visual_kinematics}
\small
\dot{{\bsm{f}}}_l^{c_\tau}\triangleq\begin{bmatrix}
\dot{u}_\tau\\\dot{v}_\tau
\end{bmatrix}=
\frac{1}{z_{l}^{c_\tau}}\cdot\bsm{A}_\tau\cdot\linvel{c}{}(\tau)+\bsm{B}_\tau\cdot\angvel{c}{}(\tau)
\end{equation}
with
\begin{equation}
\small
\bsm{A}_\tau\triangleq\begin{bmatrix}
-f_x&0&u^\prime_\tau\\
0&-f_y&v^\prime_\tau
\end{bmatrix}
\end{equation}
and
\begin{equation}
\small
\bsm{B}_\tau\triangleq\begin{bmatrix}
			\begin{aligned}
			\frac{u^\prime_\tau \cdot v^\prime_\tau}{f_y}
			\end{aligned}&
			\begin{aligned}
			-f_x-\frac{\left( u^\prime_\tau\right)^2}{f_x}
			\end{aligned}&
			\begin{aligned}
			\frac{f_x\cdot v^\prime_\tau}{f_y}
			\end{aligned}\\
			\begin{aligned}
			f_y+\frac{\left(v^\prime_\tau\right)^2}{f_y}
			\end{aligned}&
			\begin{aligned}
			-\frac{u^\prime_\tau\cdot v^\prime_\tau}{f_x}
			\end{aligned}&
			\begin{aligned}
			-\frac{f_y\cdot u^\prime_\tau}{f_x}
			\end{aligned}
			\end{bmatrix}
\end{equation}
where $\dot{{\bsm{f}}}_l^{c_\tau}$ denotes the two-dimensional pixel velocity, while $\linvel{c}{}(\tau)\triangleq\rotation{w}{c}(\tau)\cdot\linvel{c}{w}(\tau)$ and $\angvel{c}{}(\tau)\triangleq\rotation{w}{c}(\tau)\cdot\angvel{c}{w}(\tau)$ are three-dimensional camera-frame linear and angular velocities of the camera respectively;
$z_{l}^{c_\tau}$ denotes the depth of the landmark in $\coordframe{c}$ at time $\tau$;
$u^\prime_\tau\triangleq{u}_\tau-c_x$ and $v^\prime_\tau\triangleq{v}_\tau-c_y$, in which $\left(c_x,c_y \right) $ is the principal point of the camera, constituting the camera intrinsics $\bsm{x}_{\textrm{in}}^c$ together with the focal length $\left(f_x,f_y \right)$.
When multiple landmarks are tracked, both $\linvel{c}{}(\tau)$ and $\angvel{c}{}(\tau)$ can be recovered based on (\ref{equ:visual_kinematics}), see Fig. \ref{fig:visual_kinematics}.

In terms of IMU sensor model, we represent its raw inertial measurements as follows:
\begin{equation}
\small
\begin{aligned}
\tilde{\bsm{a}}=
h_a\left(\bsm{a},\bsm{x}_{\mathrm{in}}^b \right)\;&\triangleq
\bsm{a}+\bsm{b}_a+\bsm{\epsilon}_a
\\
\tilde{\bsm{\omega}}=
h_\omega\left( \bsm{\omega},\bsm{x}_{\mathrm{in}}^b \right) &\triangleq
\bsm{\omega}+\bsm{b}_\omega+\bsm{\epsilon}_\omega
\end{aligned},\;\;
\mathrm{s.t.}\;\;
\bsm{x}_{\mathrm{in}}^b\triangleq\left\lbrace \bsm{b}_a,\bsm{b}_\omega\right\rbrace
\end{equation}
where $\bsm{a}$ and $\bsm{\omega}$ denote the ideal specific force and angular velocity, while $\tilde{\bsm{a}}$ and $\tilde{\bsm{\omega}}$ are noisy and biased ones measured by the IMU;
$\bsm{b}_{(\cdot)}$ and $\bsm{\epsilon}_{(\cdot)}$ are the inertial biases and zero-mean Gaussian white noises, respectively.

\subsection{Continuous-Time State Representation}
Adhering to \emph{iKalibr} \cite{chen2024ikalibr}, the B-spline-based continuous-time state representation is employed to represent the time-varying rotation and velocity in this work.
Such a representation is more suitable for fusing asynchronous and high-frequency data compared to discrete-time ones.
Specifically, given a series of linear velocity control points:
\begin{equation}
\label{equ:vel_knots}
\small
\mathcal{V}\triangleq\left\lbrace
\bsm{v}_i,\tau_i\mid
\bsm{v}_i\in\mathbb{R}^3,\tau_i\in\mathbb{R},i\in\mathbb{N}
\right\rbrace \;\;
\mathrm{s.t.}\;\;
\tau_{i\smallplus 1}-\tau_{i}\equiv\Delta\tau_v
\end{equation}
and rotation control points:
\begin{equation}
\label{equ:rot_knots}
\small
	\mathcal{R}\triangleq\left\lbrace
	 \bsm{R}_i,\tau_i\mid
	\bsm{R}_i\in\mathrm{SO(3)},\tau_i\in\mathbb{R},i\in\mathbb{N}
	\right\rbrace\;\;
	\mathrm{s.t.}\;\;
	\tau_{i\smallplus 1}-\tau_{i}\equiv\Delta\tau_r
\end{equation}
that are temporally uniformly distributed, corresponding linear velocity $\bsm{v}(\tau)$ or rotation $\bsm{R}(\tau)$ at time $\tau\in\left[ \tau_{i},\tau_{i\smallplus 1}\right) $ in a $k$-order B-spline can be obtained as follows:
\begin{equation}
\small
\label{equ:b-spline-scale}
\bsm{v}(\tau)=\bsm{v}_i+\sum_{j=1}^{k\smallminus 1}
\lambda_j(u)
\cdot\left(\bsm{v}_{i\smallplus j}-\bsm{v}_{i\smallplus j\smallminus 1} \right) 
\end{equation}
\begin{equation}
\small
\label{equ:b-spline-rot}
\bsm{R}(\tau)=\bsm{R}_i\cdot\prod_{j=1}^{k\smallminus 1}
\mathrm{Exp}\left( \lambda_j(u)
\cdot\mathrm{Log}\left( \bsm{R}_{i\smallplus j\smallminus 1}^\top \cdot\bsm{R}_{i\smallplus j}\right) \right) 
\end{equation}
where $u=\left( \tau-\tau_{i}\right)/\Delta\tau_{s}$;
$\lambda_j(u)$ is the $j$-th element of vector $\bsm{\lambda}(u)$, obtained from the order-determined cumulative matrix and normalized time vector \cite{sommer2020efficient};
$\mathrm{Log}(\cdot)$ maps elements in the manifold of $\mathrm{SO}(3)$ to the vector space of $\mathfrak{so}(3)$, and $\mathrm{Exp}(\cdot)$ is its inverse.

\section{Methodology}
This section presents the detailed pipeline of the proposed RGBD-inertial spatiotemporal calibration.

\subsection{Overview}
\begin{figure}[t]
	\centering
	\includegraphics[width=0.98\linewidth]{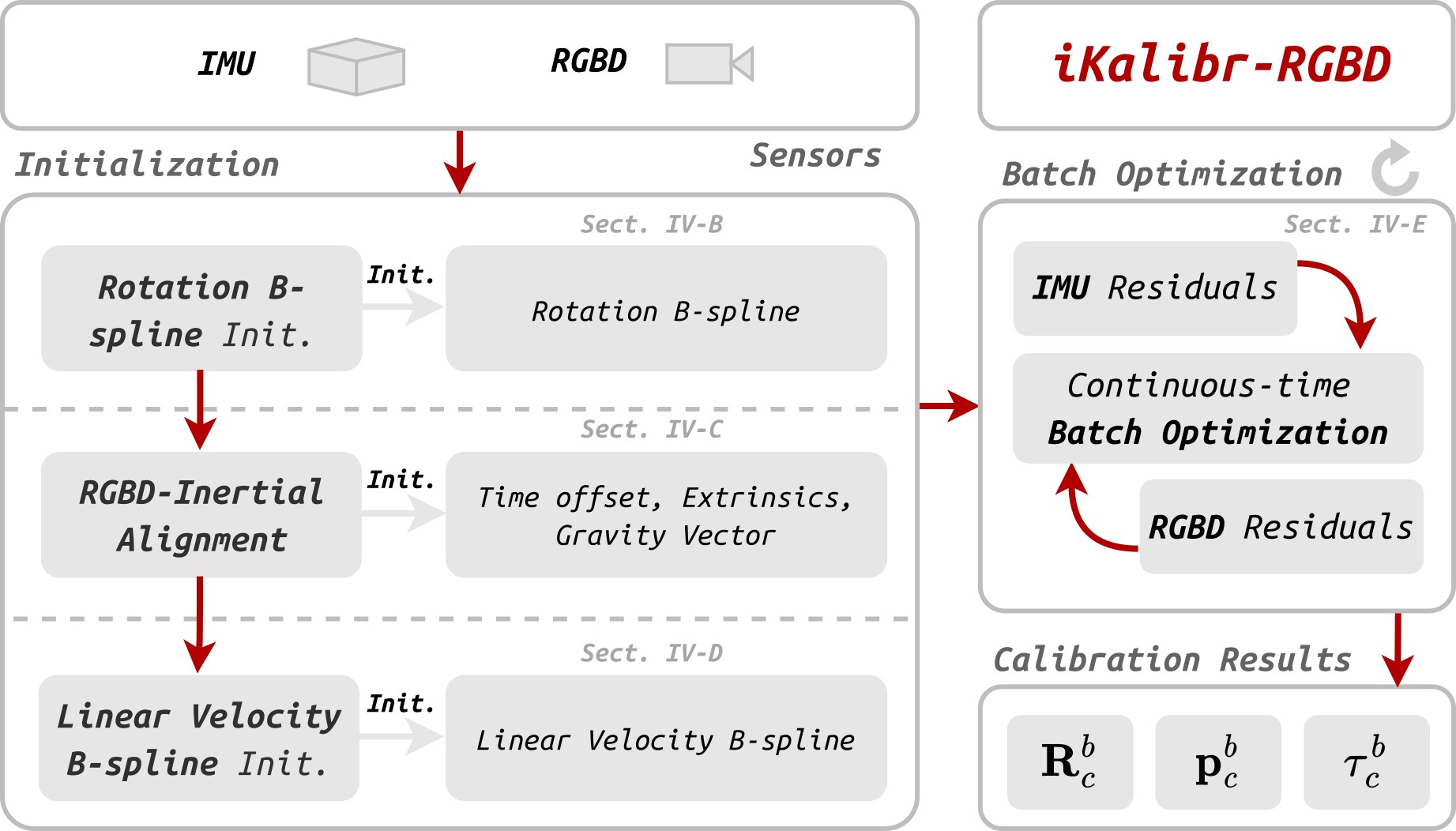}
	\caption{Illustration of the pipeline of the proposed RGBD-Inertial spatiotemporal calibration.}
	\label{fig:overview}
	\vspace{-10pt}
\end{figure}

The pipeline of the proposed RGBD-inertial spatiotemporal calibration is shown in Fig. \ref{fig:overview}.
The rotation B-spline is first recovered using raw angular velocity measurements from the gyroscope, see Section \ref{sect:rot_bspline_fit}. Following that, sparse optical flow would be conducted for ($i$) rotation-only visual odometry to initialize time offset and extrinsic rotation, and ($ii$) RGBD-only ego-velocity estimation to initialize extrinsic translation and gravity vector, see Section \ref{sect:rgbd_inertial_align}. 
RGBD-derived ego-velocities are also utilized for linear velocity B-spline recovery, see Section \ref{sect:vel_bspline_fit}.
Finally, several batch optimizations would be performed to refine all initialized states to optimal ones, see Section \ref{sect:batch_opt}.
The total state vector maintained in the estimator could be expressed as:
\begin{equation}
\small
\mathcal{X}\triangleq\left\lbrace
\rotation{c}{b},
\translation{c}{b},
\timeoffset{c}{b},
\gravity{w},
\mathcal{V},
\mathcal{R},
\bsm{x}_{\mathrm{in}}^b
\right\rbrace 
\end{equation}
where $\timeoffset{c}{b}$ denotes the time offset between the RGBD camera and IMU, satisfying $\tau_b=\tau_c+\timeoffset{c}{b}$.
Note that all control points and the gravity vector are parameterized in $\coordframe{w}$.

\subsection{Rotation B-spline Recovery}
\label{sect:rot_bspline_fit}
Based on the raw angular velocity measurements $\mathcal{D}_\omega$ from the gyroscope, the rotation B-spline could be fitted by solving the following least-squares problem:
\begin{equation}
\small
\hat{\mathcal{R}}=\arg\min\sum_{n}^{\mathcal{D}_\omega}
\left\| 
r_\omega\left(\tilde{\bsm{\omega}}_n \right) 
\right\| ^2_{\bsm{Q}_{\omega,n}}
\end{equation}
with
\begin{equation}
\label{equ:gyro_residual}
\small
\begin{aligned}
r_\omega\left(\tilde{\bsm{\omega}}_n \right) &\triangleq
h_\omega\left(\bsm{\omega}\left( \tau_n\right),\bsm{x}_{\mathrm{in}}^b \right) -\tilde{\bsm{\omega}}_n
\\
\bsm{\omega}\left( \tau\right)&=\left( \rotation{b}{w}(\tau)\right) ^\top\cdot\angvel{b}{w}(\tau)
\end{aligned}
\end{equation}
where $\rotation{b}{w}(\tau)$ and $\angvel{b}{w}(\tau)$ are the rotation and angular velocity obtained from the rotation B-spline based on (\ref{equ:b-spline-rot}), which exactly introduces the estimation of rotational control points $\mathcal{R}$;
$\bsm{Q}_{\omega,n}$ denotes the information matrix, determined by the noise level of the $n$-th angular velocity measurement $\tilde{\bsm{\omega}}_n$ at time $\tau_n$.

\subsection{RGBD-Inertial Alignment}
\label{sect:rgbd_inertial_align}
The KLT-based sparse optical flow \cite{lucas1981iterative} is first performed in this stage to track stable features for rotation-only visual odometry and subsequent RGBD-only ego-velocity estimation. 

\subsubsection{Rotation-Only Visual Odometry}
Based on the tracked features, the frame-to-frame direct
rotation optimization \cite{kneip2013direct} would be employed to estimate the relative rotations between two consecutive frames, where the random sample consensus (RANSAC) is employed simultaneously to reject outliers in tracked feature pairs.
Subsequently, the rotation-only hand-eye alignment would be conducted to recover the extrinsic rotation and time offset, which can be expressed as:
\begin{equation}
\small
\label{equ:rot_hand_eye}
\left\lbrace \rotationhat{c}{b},\timeoffsethat{c}{b}\right\rbrace =\arg\min\sum^{\mathcal{S}_{c}}_n
\left\| 
\rotationhat{c}{b}\cdot\rotation{c_{n\smallplus 1}}{c_{n}}
\left( \rotation{b_{n\smallplus 1}}{b_{n}}
\rotationhat{c}{b}\right) ^\top
\right\| ^2_{\bsm{Q}_{c,n}}
\end{equation}
with
\begin{equation}
\small
\rotation{b_{n\smallplus 1}}{b_{n}}\triangleq
\left( \rotation{b}{w}(\tau_{n}+\timeoffsethat{c}{b})\right) ^\top
\cdot\rotation{b}{w}(\tau_{n\smallplus 1}+\timeoffsethat{c}{b})
\end{equation}
where $\rotation{c_{n\smallplus 1}}{c_{n}}\in\mathcal{S}_{c}$ is the estimated frame-to-frame rotation; $\rotation{b}{w}(\cdot)$ denotes the time-varying rotation of the IMU, queried from the fitted rotation B-spline.

\subsubsection{RGBD-Only Ego-Velocity Estimation}
\begin{figure}[t]
	\centering
	\begin{minipage}{\linewidth}
		\centerline{\includegraphics[width=\textwidth]{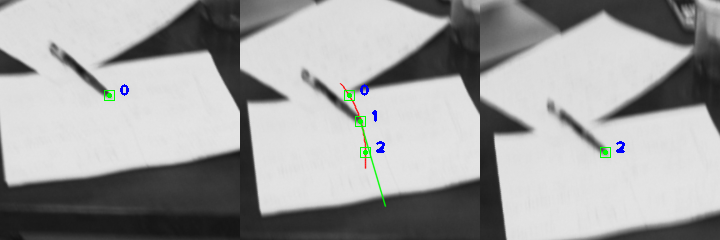}}
		\centerline{\includegraphics[width=\linewidth]{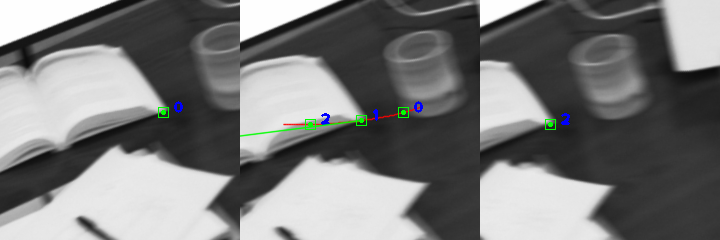}}
	\end{minipage}
	\caption{Schematic of two-dimensional pixel velocities computed using three-point first-order Lagrange polynomial. Red solid curve represents the feature trajectory from the (zero-order) Lagrange polynomial, while green line is the computed pixel velocity of the middle point (the feature indexed as $1$).}
	\label{fig:pixel_vel}
\end{figure}

Using the outlier-rejected tracked features and available depth information from the RGBD camera, the ego-velocity of the camera at time $\tau$, i.e., $\linvel{c}{}(\tau)$, can be obtained by solving the following linear least-squares problem based on (\ref{equ:visual_kinematics}):
\begin{equation}
\label{equ:rgbd_only_vel}
\small
\linvelhat{c}{}(\tau)=\arg\min\sum_{m}^{\mathcal{S}_{c_\tau}}
\left\| 
\frac{\bsm{A}_{\tau,m}\cdot\linvelhat{c}{}(\tau)}{z_{l_m}^{c_\tau}}+\bsm{B}_{\tau,m}\cdot\angvel{c}{}(\tau)-\dot{{\bsm{f}}}_{l_m}^{c_\tau}
\right\| ^2_{\bsm{Q}_{c,m}}
\end{equation}
with
\begin{equation}
\small
\angvel{c}{}(\tau)=\left( \rotation{b}{w}(\tau+\timeoffset{c}{b})\cdot
\rotation{c}{b}\right) ^\top\cdot\angvel{b}{w}(\tau+\timeoffset{c}{b})
\end{equation}
where $\mathcal{S}_{c_\tau}$ denotes the tracked feature sequence in current frame;
camera-frame angular velocity $\angvel{c}{}(\tau)$ are pre-computed based on the fitted B-spline, extrinsic rotation, and time offset;
$\dot{{\bsm{f}}}_{l_m}^{c_\tau}$ is the pixel velocity of the feature on the image plane, which could be obtained by numerical differentiation.
In this work, the three-point first-order Lagrange polynomial \cite{werner1984polynomial} is utilized, which is defined as follows by considering $\tau$ as the independent variable and $y\simeq u/v$ as the dependent variable:
\begin{equation}
\label{equ:three_point_lp}
\small
\dot{y}(\tau)=\sum_{i=0}^{n}\left( y_i\cdot
 \left( \prod_{j=0,j\ne i}^{n}\frac{\tau-\tau_j}{\tau_i-\tau_j}\right) \cdot\left(
\sum_{j=0,j\ne i}^{n}\frac{1}{\tau-\tau_j}\right) \right) 
\end{equation}
where $n\equiv 3$. Each component of $\dot{{\bsm{f}}}_{l_m}^{c_\tau}$ can be independently computed based on (\ref{equ:three_point_lp}), if it's tracked at least three consecutive frames, see Fig. \ref{fig:pixel_vel}.

\subsubsection{Extrinsic Translation and Gravity Recovery}
Based on the estimated camera-frame ego-velocity sequence $\mathcal{S}_{\mathrm{vel}}$, the extrinsic translation can be recovered by aligning specific force measurements with $\mathcal{S}_{\mathrm{vel}}$, which could simultaneously recover the world-frame gravity.
This can be described as the following least-squares problem:
\begin{equation}
\small
\left\lbrace\translationhat{c}{b}, \hat{\bsm{g}}^w\right\rbrace= 
\arg\min\sum_{n}^{\mathcal{S}_{\mathrm{vel}}}\left\| 
\linvel{b}{w}(\tau_{n\smallplus 1})-\linvel{b}{w}(\tau_{n})-\Delta\linvel{n,n\smallplus 1}{b}
\right\| ^2_{\bsm{Q}_{c,n}}
\end{equation}
with
\begin{equation}
\label{equ:body_vel_from_cam_vel}
\small
\begin{aligned}
\linvel{b}{w}(\tau)&=
\rotation{b}{w}(s)\cdot\rotation{c}{b}\cdot\linveltilde{c}{}(\tau)-
\liehat{\angvel{b}{w}(s)}\cdot\rotation{b}{w}(s)\cdot\translationhat{c}{b}
\\
\Delta\linvel{n,n\smallplus 1}{b}&\triangleq
\int_{\tau_n}^{\tau_{n\smallplus 1}}\rotation{b}{w}(s^\prime)\cdot\bsm{a}(s^\prime)\cdot\mathrm{d}t
+\gravityhat{w}\cdot\Delta\tau_{n,n\smallplus 1}
\\
s&\triangleq\tau+\timeoffset{c}{b},\;\;
s^\prime\triangleq t+\timeoffset{c}{b},\;\;
\Delta\tau_{n,n\smallplus 1}\triangleq\tau_{n\smallplus 1}-\tau_{n}
\end{aligned}
\end{equation}
where $\linvel{b}{w}(\tau)$ is the linear velocity of the IMU, transferred from the estimated RGBD-derived ego-velocity $\linveltilde{c}{}(\tau)\in\mathcal{S}_{\mathrm{vel}}$;
$\liehat{\cdot}$ represents the cross-product operation;
$\Delta\linvel{n,n\smallplus 1}{b}$ denotes the linear velocity variation of the IMU in timepiece $\left[\tau_{n},\tau_{n\smallplus 1} \right)$, computed through numerical integration on discrete specific force measurements.

\subsection{Linear Velocity B-spline Recovery}
\label{sect:vel_bspline_fit}
Subsequently, the linear velocity B-spline could be initialized based on the fitted rotation B-spline, recovered spatiotemporal parameters, and RGBD-derived ego-velocities:
\begin{equation}
\small
\hat{\mathcal{V}}=\arg\min\sum_{n}^{\mathcal{S_{\mathrm{vel}}}}\left\|\linvel{b}{w}(\tau)-\linveltilde{b}{w}(\tau) \right\| ^2_{\bsm{Q}_{n}}
\end{equation}
where $\linvel{b}{w}(\tau)$ is obtained based on (\ref{equ:b-spline-scale}), which introduces the optimization of linear velocity control points; $\linveltilde{b}{w}(\tau)$ is pre-computed linear velocity using RGBD-derived ego-velocities and already-initialized states:
\begin{equation}
\small
\linveltilde{b}{w}(\tau)=
\rotation{b}{w}(s)\cdot\rotation{c}{b}\cdot\linveltilde{c}{}(\tau)-
\liehat{\angvel{b}{w}(s)}\cdot\rotation{b}{w}(s)\cdot\translation{c}{b}
\end{equation}
which is exactly the same form of (\ref{equ:body_vel_from_cam_vel}).
The only difference is that the extrinsic translation $\translation{c}{b}$ is no longer a quantity to be estimated here.
At this point, the multi-stage initialization is complete, where all spatiotemporal parameters and other interested ones in the estimator are recovered rigorously.
As for inertial biases, they are set to zero vectors and would be estimated in the final batch optimization.

\subsection{Continuous-Time Batch Optimization}
\label{sect:batch_opt}
Finally, a continuous-time factor graph would be organized and solved to refine all initialized states to optimal ones, into which raw inertial measurements and visual optical flow are tightly coupled.

\subsubsection{Inertial Residuals}
The raw angular velocity and specific force measurements would be utilized in gyroscope and accelerometer residuals, respectively.
Specifically, the gyroscope residual is exactly the one in (\ref{equ:gyro_residual}), denoted as $r_\omega\left(\tilde{\bsm{\omega}}_n \right)$, which had been utilized for rotation B-spline recovery.
As for the accelerometer residual $r_a\left(\tilde{\bsm{a}}_n \right)$, it's expressed as follows:
\begin{equation}
\small
r_{a}\left( \tilde{\bsm{a}}_n\right) \triangleq
h_a\left( {\bsm{a}}(\tau_n),\bsm{x}_{\mathrm{in}}^{b}\right) 
-\tilde{\bsm{a}}_n
\end{equation}
with
\begin{equation}
\small
{\bsm{a}}(\tau)=
\left( \rotation{b}{w}(\tau) \right)^\top\cdot
\left(\linacce{b}{w}(\tau) -\gravityhat{w}\right) 
\end{equation}
which introduces the optimization of the gravity vector, as well as control points of both rotation and linear velocity B-splines by $\rotation{b}{w}(\tau)$ and $\linacce{b}{w}(\tau)$.

\subsubsection{Visual Optical Flow Residual}
The visual residual is similar to the one utilized in (\ref{equ:rgbd_only_vel}). The only difference is that the camera-frame ego-velocities are obtained from the B-splines, rather than pre-computed ones from RGBD-only ego-velocity estimation, to introduce the optimization of control points.
The visual residual $r_{c}\left( 
\dot{{\bsm{f}}}_{l_m}^{c_n}
\right)$ is expressed as follows:
\begin{equation}
\small
r_{c}\left( 
\dot{{\bsm{f}}}_{l_m}^{c_n}
\right)\triangleq
\frac{1}{z_{l_m}^{c_n}}\cdot\bsm{A}_{\tau,m}\cdot\linvel{c}{}(\tau+\timeoffsethat{c}{b})+\bsm{B}_{\tau,m}\cdot\angvel{c}{}(\tau+\timeoffsethat{c}{b})-\dot{{\bsm{f}}}_{l_m}^{c_n}
\end{equation}
with
\begin{equation}
\small
\begin{aligned}
\linvel{c}{}(\tau)&=\left(\rotation{b}{w}(\tau)\cdot\rotationhat{c}{b} \right)^\top\cdot\left(\linvel{b}{w}(\tau)-
\liehat{\rotation{b}{w}(\tau)\cdot\translationhat{c}{b}}
 \right) \cdot\angvel{b}{w}(\tau)
 \\
 \angvel{c}{}(\tau)&=\left(\rotation{b}{w}(\tau)\cdot\rotationhat{c}{b} \right)^\top\cdot\angvel{b}{w}(\tau)
\end{aligned}.
\end{equation}

\subsubsection{Factor Graph Optimization}
The final continuous-time factor graph can be described as the following non-linear least-squares problem:
\begin{equation}
\label{equ:bo}
\small
\begin{gathered}
\hat{\mathcal{X}}=\arg\min\sum_{n}^{\mathcal{D}_{\omega}}
\rho_\omega\left(\left\| 
r_{\omega}\left( \tilde{\bsm{\omega}}_n\right) 
\right\| ^2_{\bsm{Q}_{\omega,n}}\right) +
\\
\sum_{n}^{\mathcal{D}_{a}}
\rho_a\left(\left\| 
r_{a}\left( \tilde{\bsm{a}}_n\right) 
\right\| ^2_{\bsm{Q}_{a,n}}\right) 
+
\sum_{n}^{\mathcal{D}_{c}}\sum_{m}^{\mathcal{S}_{c_n}}
\rho_c\left(\left\| 
r_{c}\left( \dot{{\bsm{f}}}_{l_m}^{c_n}\right)
\right\| ^2_{\bsm{Q}_{v,n,m}}\right) 
\end{gathered}
\end{equation}
where $\rho(\cdot)$ are the Huber loss functions; $\mathcal{D}_w$ and $\mathcal{D}_a$ denote the inertial measurement sets, while $\mathcal{D}_c$ is frame set from the RGBD camera; $\mathcal{S}_{c_n}$ represents the tracked feature sequence in the $n$-th camera frame.
The non-linear problem (\ref{equ:bo}) would be optimized several times using \emph{Ceres solver} \cite{Agarwal_Ceres_Solver_2022} until the final convergence.

\section{Real-World Experiment}
\begin{figure}[t]
	\centering
	\includegraphics[width=\linewidth]{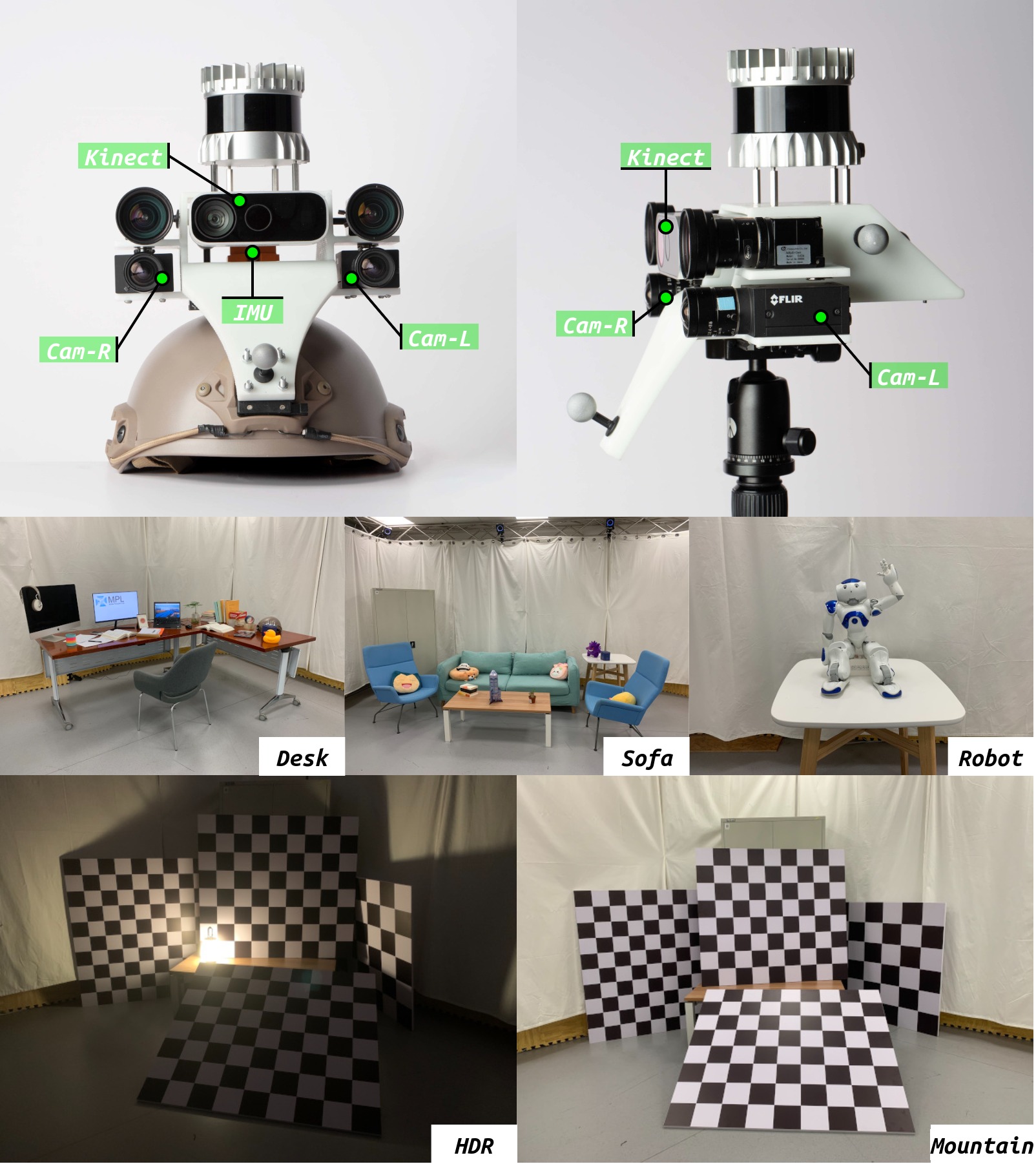}
	\caption{The sensor suite and five scenarios in the real-world experiments. Datasets are from the open-source \emph{VECtor Benchmark} \cite{gao2022vector}.}
	\label{fig:sensor_suite}
\end{figure}

A sufficient number of real-world experiments were
carried out to verify the feasibility of the proposed \emph{iKalibr-RGBD} and evaluate its performance in terms of calibration repeatability, accuracy, consistency, and computation consumption.

\subsection{Dataset}
The open-source \emph{VECtor Benchmark} \cite{gao2022vector} is utilized in real-world experiments to strengthen the reliability of the evaluation.
Five data sequences in \emph{VECtor Benchmark} with sufficiently excited motions are considered to ensure the observability of spatiotemporal parameters, see Fig. \ref{fig:sensor_suite}.
While several kinds of sensors are integrated in the suite, data streams from the \emph{Azure Kinect}, two \emph{FLIR Grasshopper3} cameras, and \emph{XSens MTi-30 AHRS} are involved in calibration.
The acquisition rate for cameras is set to 30Hz, and for IMU, that is 200Hz.
Note that depth images from the \emph{Azure Kinect} sensor were pre-aligned to coordinate frames of two \emph{FLIR} cameras using the attached program
 from \emph{VECtor Benchmark}, to construct RGBD data stream.

\subsection{Spatiotemporal Convergence}
\begin{figure}[t]
	\centering
	\includegraphics[width=\linewidth]{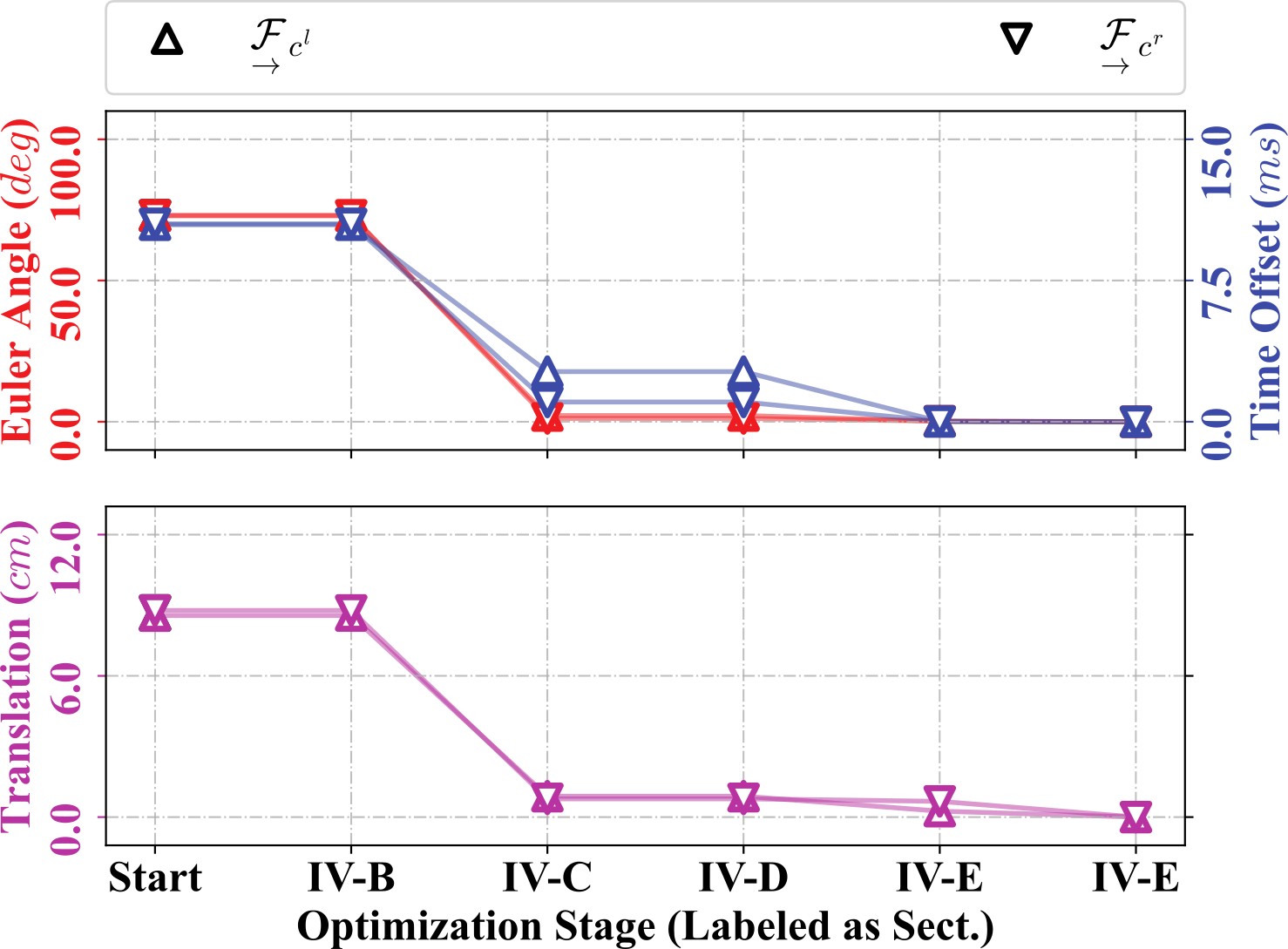}
	\caption{The convergence performance of \emph{iKalibr-RGBD}. Quantities are the RMSEs of estimates with respect to the final ones. Horizontal labels indicate calibration stages and are matched with section titles in this paper.}
	\label{fig:convergence}
	\vspace{-10pt}
\end{figure}

The spatiotemporal convergence performance of the proposed \emph{iKalibr-RGBD} is shown in Fig. \ref{fig:convergence}, where spatiotemporal parameters of two cameras are with respect to the IMU.
All spatiotemporal parameters at the beginning are set to zeros or identities.
For a better understanding by readers, the horizontal axis is labeled by indexes of corresponding sections.
As can be seen, in the initialization procedure (Section \ref{sect:rot_bspline_fit}, \ref{sect:rgbd_inertial_align}, and \ref{sect:vel_bspline_fit}), the extrinsic rotation, time offset, and extrinsic translation are initialized sequentially, where two B-splines are also recovered.
Subsequently, two batch optimizations are conducted (Section \ref{sect:batch_opt}), which refines all states to the global optimal ones.
These results demonstrate that \emph{iKalibr-RGBD} could effectively and efficiently estimate spatiotemporal parameters and reach the final convergence.

\subsection{Accuracy and Repeatability Comparison}
\begin{table}[t]
\centering
\caption{Spatiotemporal Calibration Results in Experiments}
\label{tab:calib_results}
\begin{threeparttable}
\begin{tabular}{cc|rrrr}
\toprule
\multicolumn{2}{c|}{Param.}                                                             & \multicolumn{1}{c}{iKalibr-RGBD} & \multicolumn{1}{c}{iKalibr \cite{chen2024ikalibr}} & \multicolumn{1}{l}{Kalibr \cite{furgale2013unified}} & \multicolumn{1}{c}{CAD}       \\ \midrule\midrule
\multicolumn{1}{c|}{\multirow{8}{*}{\rotatebox{90}{Cam-L To IMU}}}   & $\hat{p}_{x}$    & 12.27$\pm$0.47                   & 12.20$\pm$0.42                                     & 14.16\;\;                                            & 11.32\;\;                     \\
\multicolumn{1}{c|}{}                                                & $\hat{p}_{y}$    & 8.32$\pm$0.22                    & 8.35$\pm$0.11                                      & 7.36\;\;                                             & 8.25\;\;                      \\
\multicolumn{1}{c|}{}                                                & $\hat{p}_{z}$    & -1.79$\pm$0.23                   & -1.50$\pm$0.14                                     & -1.47\;\;                                            & -1.45\;\;                     \\ \cmidrule{2-6} 
\multicolumn{1}{c|}{}                                                & $\hat{\theta}_r$ & 90.54$\pm$0.05                   & 90.55$\pm$0.03                                     & 90.97\;\;                                            & 90.00\;\;                     \\
\multicolumn{1}{c|}{}                                                & $\hat{\theta}_p$ & -179.51$\pm$0.02                 & -179.51$\pm$0.01                                   & -179.54\;\;                                          & -180.00\;\;                   \\
\multicolumn{1}{c|}{}                                                & $\hat{\theta}_y$ & 90.83$\pm$0.08                   & 90.07$\pm$0.02                                     & 90.46\;\;                                            & 90.00\;\;                     \\ \cmidrule{2-6} 
\multicolumn{1}{c|}{}                                                & $\hat{\tau}$     & 9.62$\pm$0.80                    & 9.66$\pm$0.70                                      & 8.46\;\;                                             & \multicolumn{1}{c}{\ding{55}} \\ \midrule
\multicolumn{1}{c|}{\multirow{8}{*}{\rotatebox{90}{Cam-R To IMU}}}   & $\hat{p}_{x}$    & 12.34$\pm$0.41                   & 12.30$\pm$0.46                                     & 13.97\;\;                                            & 11.32\;\;                     \\
\multicolumn{1}{c|}{}                                                & $\hat{p}_{y}$    & -8.48$\pm$0.24                   & -8.72$\pm$0.11                                     & -9.77\;\;                                            & -8.75\;\;                     \\
\multicolumn{1}{c|}{}                                                & $\hat{p}_{z}$    & -1.71$\pm$0.30                   & -1.42$\pm$0.21                                     & -1.37\;\;                                            & -1.45\;\;                     \\ \cmidrule{2-6} 
\multicolumn{1}{c|}{}                                                & $\hat{\theta}_r$ & 90.30$\pm$0.02                   & 90.25$\pm$0.05                                     & 90.56\;\;                                            & 90.00\;\;                     \\
\multicolumn{1}{c|}{}                                                & $\hat{\theta}_p$ & 179.05$\pm$0.03                  & 179.05$\pm$0.02                                    & 179.16\;\;                                           & -180.00\;\;                   \\
\multicolumn{1}{c|}{}                                                & $\hat{\theta}_y$ & 90.59$\pm$0.09                   & 90.27$\pm$0.02                                     & 90.36\;\;                                            & 90.00\;\;                     \\ \cmidrule{2-6} 
\multicolumn{1}{c|}{}                                                & $\hat{\tau}$     & 9.61$\pm$0.72                    & 9.64$\pm$0.70                                      & 8.39\;\;                                             & \multicolumn{1}{c}{\ding{55}} \\ \midrule
\multicolumn{1}{c|}{\multirow{8}{*}{\rotatebox{90}{Cam-R To Cam-L}}} & $\hat{p}_{x}$    & 16.80$\pm$0.26                   & 17.07$\pm$0.06                                     & 17.14\;\;                                            & 17.00\;\;                     \\
\multicolumn{1}{c|}{}                                                & $\hat{p}_{y}$    & 0.06$\pm$0.12                    & 0.05$\pm$0.09                                      & 0.03\;\;                                             & 0.00\;\;                      \\
\multicolumn{1}{c|}{}                                                & $\hat{p}_{z}$    & -0.19$\pm$0.29                   & -0.17$\pm$0.07                                     & -0.48\;\;                                            & 0.00\;\;                      \\
\multicolumn{1}{c|}{}                                                & $\hat{\theta}_r$ & -0.22$\pm$0.05                   & -0.30$\pm$0.04                                     & -0.09\;\;                                            & 0.00\;\;                      \\
\multicolumn{1}{c|}{}                                                & $\hat{\theta}_p$ & 0.35$\pm$0.08                    & 0.20$\pm$0.03                                      & 0.42\;\;                                             & 0.00\;\;                      \\
\multicolumn{1}{c|}{}                                                & $\hat{\theta}_y$ & 1.43$\pm$0.02                    & 1.43$\pm$0.02                                      & 1.29\;\;                                             & 0.00\;\;                      \\ \cmidrule{2-6} 
\multicolumn{1}{c|}{}                                                & $\hat{\tau}$     & -0.01$\pm$0.04                   & -0.01$\pm$0.02                                     & -0.07\;\;                                            & \multicolumn{1}{c}{\ding{55}} \\ \bottomrule
\end{tabular}
\begin{tablenotes} 
\item[*] Extrinsic translations in $(cm)$, extrinsic Euler angles in $(deg)$, and time offsets in $(ms)$.
\item[*] Results from \emph{Kalibr} and \emph{CAD} are provided by \emph{VECtor Benchmark} \cite{gao2022vector}.
\end{tablenotes}
\end{threeparttable}
\vspace{-10pt}
\end{table}
The calibration accuracy and repeatability of \emph{iKalibr-RGBD} are subsequently evaluated. 
Table \ref{tab:calib_results} summarizes the calibration results of \emph{iKalibr-RGBD} in the five scenarios mentioned above, where spatiotemporal parameters from \emph{Kalibr} \cite{furgale2013unified} and computer aided design (CAD) provided by \emph{VECtor Benchmark} \cite{gao2022vector} are also listed.
To evaluate the repeatability, the visual-inertial calibration module of \emph{iKalibr} \cite{chen2024ikalibr} orienting to optical cameras is also considered in the experiments for a reliable comparison.
It could be found that the proposed \emph{iKalibr-RGBD} achieves comparable calibration results with its predecessor \emph{iKalibr}.
The STDs of extrinsic rotation, extrinsic translation, and time offset are within $0.1\;deg$, $0.5\;cm$, and $0.8\;ms$, respectively.
Although \emph{iKalibr-RGBD} performs slightly worse than \emph{iKalibr} and \emph{Kalibr} in terms of repeatability (see the STDs) and accuracy, mainly on scale-related extrinsic translation, due to employing non-primitive visual observations and the lack of constraints from explicit priori geometric scale compared with \emph{Kalibr}, it has obvious advantages in terms of computation efficiency, see Section \ref{sect:comp_consumption} and Table \ref{tab:comp_consumption}.

\subsection{Consistency Evaluation}

\begin{figure}[t]
	\centering
	\includegraphics[width=0.9\linewidth]{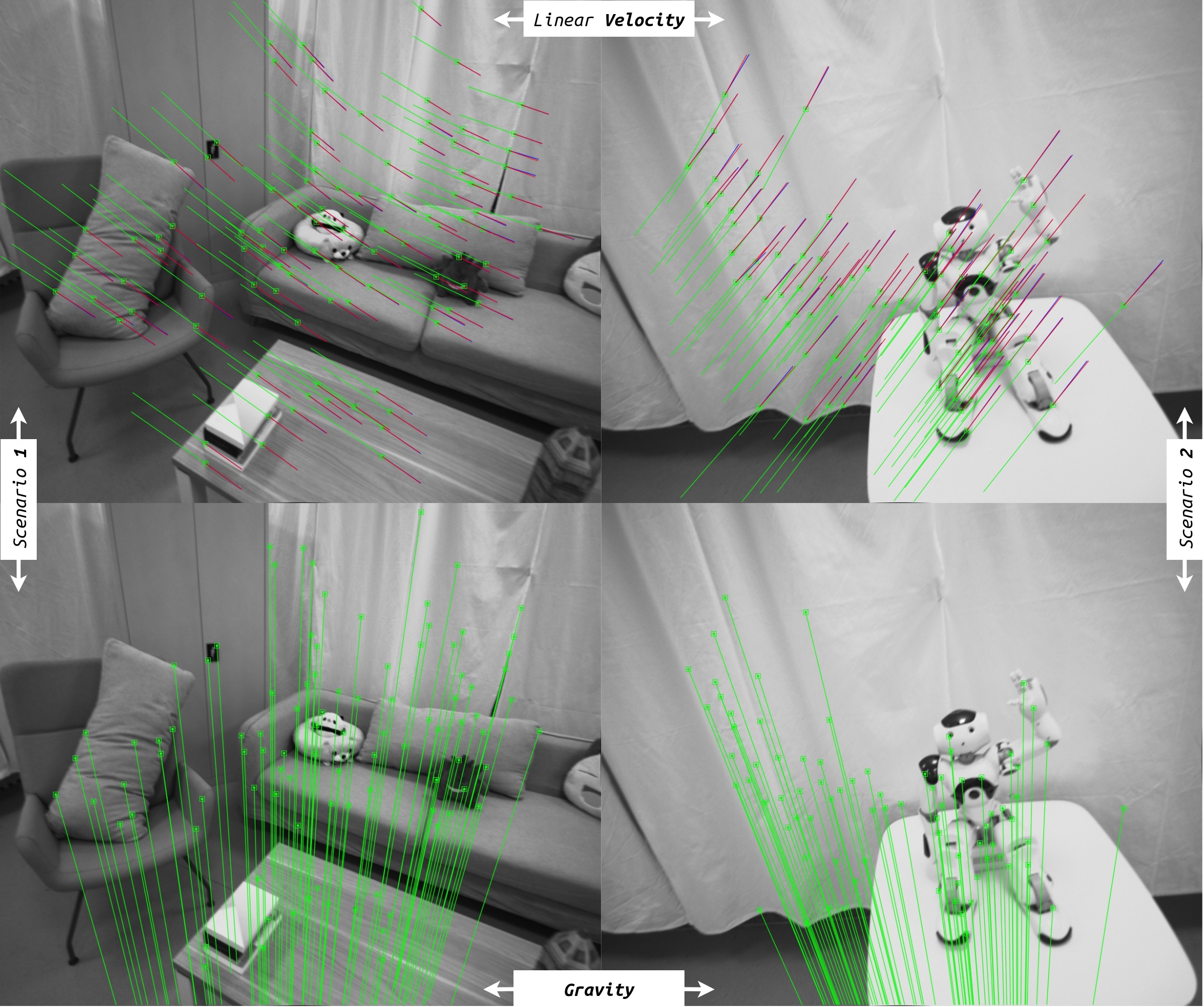}
	\caption{Visual kinematics of different moments from two scenarios, where the top row indicates linear velocities (green lines) and the bottom row represents gravity vectors (green lines). Red and blue solid lines in the top images are (scaled) estimates-derived and optical-flow-derived pixel velocities.}
	\label{fig:consistency}
\end{figure}

\begin{figure}[t]
	\centering
	\includegraphics[width=0.9\linewidth]{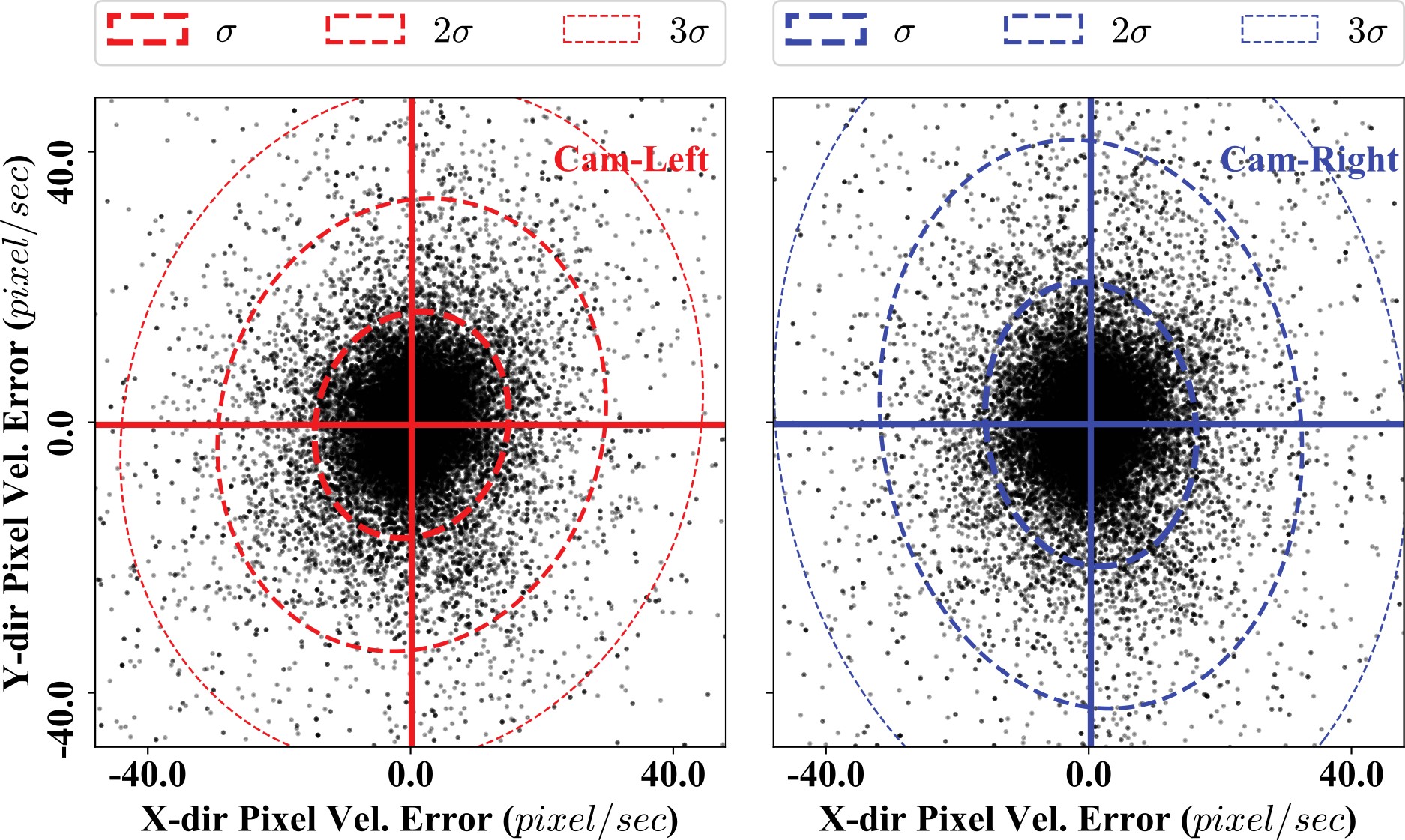}
	\caption{The distribution of visual optical flow residuals for two RGBDs after final convergence. Solid lines are means of residuals, while circles represent the variance.}
	\label{fig:rgbd_vel_error}
	\vspace{-10pt}
\end{figure}

The calibration consistency is subsequently evaluated based on the by-products from spatiotemporal determination.
Based on the estimated B-splines, we project the world-frame gravity vector and linear velocities of tracked landmarks with respect to the camera onto the image plane, see Fig. \ref{fig:consistency}.
Estimates-derived and optical-flow-derived pixel velocities of features, i.e., $\hat{\dot{\bsm{f}}}_{l_m}^{c_n}$ from (\ref{equ:visual_kinematics}) and $\tilde{\dot{\bsm{f}}}_{l_m}^{c_n}$ from (\ref{equ:three_point_lp}), are also plotted to qualitatively evaluate consistency.
As can be seen, the direction of estimated gravity is consistent well with the structures of surroundings.
Meanwhile, estimates-derived pixel velocities are well matched with the ones derived by optical flow.
To quantitatively evaluate their difference (i.e., visual residual),
the two-dimensional visual optical flow residuals $r_{c}\left( 
\dot{{\bsm{f}}}_{l_m}^{c_n}
\right)$ after final spatiotemporal convergence are statistically analyzed and depicted in Fig. \ref{fig:rgbd_vel_error}.
It could be found that each dimension of errors obeys zero-mean Gaussian distribution, and STDs are generally within $20\;pixels/sec$.
These results show the calibration consistency \emph{iKalibr-RGBD} is capable of.

\subsection{Computation Consumption Comparison}
\label{sect:comp_consumption}

\begin{table}[t]
\centering
\caption{Computation Consumption Comparison in Experiments}
\label{tab:comp_consumption}
\begin{threeparttable}
\begin{tabular}{c|cclll}
\toprule
\multirow{4}{*}{Config.} & OS Name                      & \multicolumn{4}{l}{Ubuntu 20.04.6 LTS 64-Bit} \\ \cmidrule{2-6} 
                         & Processor                    & \multicolumn{4}{l}{12th Gen Intel® Core™ i9}  \\ \cmidrule{2-6} 
                         & Graphics                     & \multicolumn{4}{l}{Mesa Intel® Graphics}      \\ \midrule\midrule
\multirow{2.5}{*}{Scenes}  & \multicolumn{5}{c}{Computation Consumption (\textbf{unit: minute})}          \\ \cmidrule{2-6} 
                         & iKalibr-RGBD (Init.+BOs)     & \multicolumn{4}{c}{iKalibr (Init.+SfM+BOs)}   \\ \midrule
Desk                     & 0.620+0.248                  & \multicolumn{4}{c}{0.165+37.875+0.586}        \\
Sofa                     & 0.675+0.239                  & \multicolumn{4}{c}{0.158+32.101+0.629}        \\
Robot                    & 0.608+0.201                  & \multicolumn{4}{c}{0.172+43.230+0.601}        \\
HDR                      & 0.584+0.191                  & \multicolumn{4}{c}{0.175+30.037+0.532}        \\
Mountain                 & 0.653+0.209                  & \multicolumn{4}{c}{0.156+31.691+0.551}        \\ \midrule
Avg.                     & 0.846                        & \multicolumn{4}{c}{35.732}                    \\ \midrule\midrule
Kalibr                   & Corner Extraction+Opt. & \multicolumn{4}{c}{0.252+1.272=1.524}         \\ \bottomrule
\end{tabular}
\begin{tablenotes} 
\item[*] Time consumption statistics are based on spatiotemporal calibration of a single-camera (left camera) single-IMU sensor configuration. Data used in calibration lasts $30\;sec$ (about 900 images).
\item[*] All other variables that have a large impact on time consumption in calibration, e.g., the time distance of B-splines $\Delta\tau_v$ in (\ref{equ:vel_knots}) and $\Delta\tau_r$ in (\ref{equ:rot_knots}), were set to the same value in the experiment.
\item[*] \emph{BOs} means the conducted several continuous-time batch optimizations.
\end{tablenotes}
\end{threeparttable}
\vspace{-10pt}
\end{table}

The most impressive point of the proposed \emph{iKalibr-RGBD} is its efficient computation, compared with the visual-inertial calibration of its predecessor \emph{iKalibr} and \emph{Kalibr}, see Table \ref{tab:comp_consumption}.
It can be found that \emph{iKalibr-RGBD} outperforms \emph{iKalibr} and \emph{Kalibr} on computational efficiency.
Such a high performance mainly comes from the mapping-free ego-velocity estimation \emph{iKalibr-RGBD} employed, which does not require time-consuming SfM \cite{schoenberger2016sfm} for visual-inertial initialization and subsequent mapping-based continuous-time trajectory optimization \emph{iKalibr} adopted.
Meanwhile, as the image acquisition frequency increases, such a difference in computational consumption will be more significant.

\section{Conclusion}
In this paper, we present a targetless continuous-time visual-inertial spatiotemporal calibration method for RGBDs, termed as \emph{iKalibr-RGBD}, a follow-up work of \emph{iKalibr} \cite{chen2024ikalibr}.
Considering additional available depth information from RGBD camera, we remodel the visual residual based on optical flow, and employ the mapping-free ego-velocity estimation, instead of the previous mapping-based trajectory estimation, for low-computation spatiotemporal calibration.
The proposed calibration method starts with a rigorous multi-stage initialization procedure, which sequentially recovers the rotation B-spline, spatiotemporal parameters, gravity vector, and linear velocity B-spline.
Subsequently, a continuous-time-based factor graph would be organized and solved to refine all states in the estimator to global optimal ones.
Adequate real-world experiments were conducted to validate the effectiveness of the proposed method and evaluate its performance.
The experimental results show that \emph{iKalibr-RGBD} achieves comparable results with its predecessor \emph{iKalibr} in terms of visual-inertial calibration, and has significant advantages in computation efficiency.
In future work, we will further the accuracy and repeatability of \emph{iKalibr-RGBD} to catch up with visual-inertial calibration in \emph{iKalibr}, and focus on supports of other sensor types into \emph{iKalibr}.

\section*{Acknowledgments}
The optimization for code implementation of \emph{iKalibr-RGBD} was assisted partially by the Supercomputing Center and the GREAT Group of Wuhan University.

\section*{CRediT Authorship Contribution Statement}
\textbf{Shuolong Chen}: Conceptualisation, Methodology, Software, Validation, Original Draft.
\textbf{Xingxing Li}: Supervision, Funding Acquisition, Review and Editing.
\textbf{Shengyu Li} and \textbf{Yuxuan Zhou}: Review and Editing.
	
\bibliographystyle{IEEEtran}
\bibliography{reference}
	
\newcommand{\vspacebio}{\vspace{-1 cm}}
\vspacebio
\begin{IEEEbiography}[{\includegraphics[width=1in,height=1.25in,clip,keepaspectratio,cframe={black!8!white}]{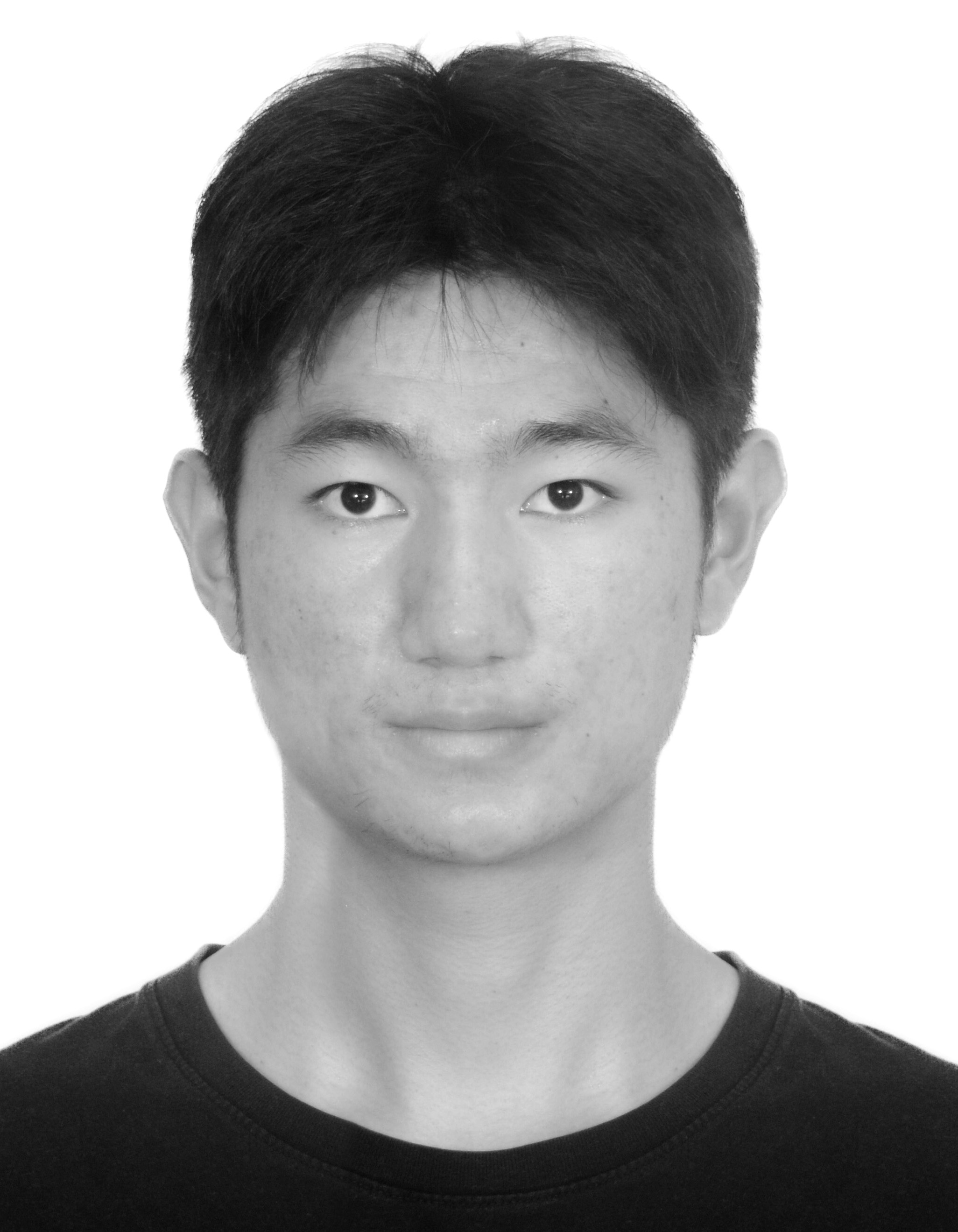}}]{Shuolong Chen}
	received the B.S. degree in geodesy and geomatics engineering from Wuhan University, Wuhan China, in 2023.
	
	He is currently a master candidate at the School of Geodesy and Geomatics (SGG), Wuhan University. His area of research currently focuses on integrated navigation systems and multi-sensor fusion, mainly on spatiotemporal calibration.
	
	Contact him via	e-mail: \emph{shlchen@whu.edu.cn}
\end{IEEEbiography}
\vspacebio
\begin{IEEEbiography}[{\includegraphics[width=1in,height=1.25in,clip,keepaspectratio,cframe={black!8!white}]{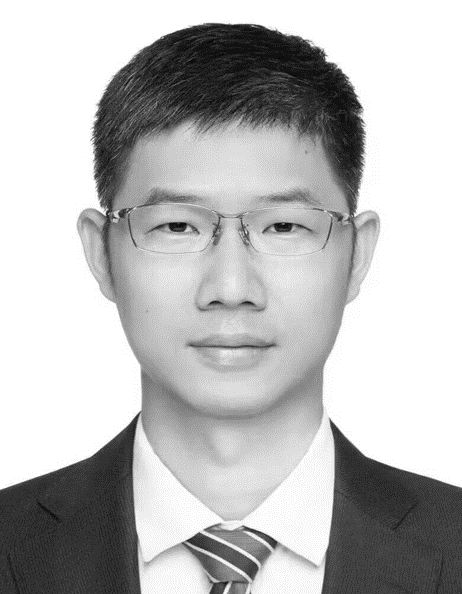}}]{Xingxing Li}
	received the B.S. degree from Wuhan University, Wuhan China, in 2008, and the Ph.D. degree from the Department of Geodesy and Remote Sensing, German Research Centre for Geosciences (GFZ), Potsdam, Germany, in 2015.
	
	He is currently a professor at the Wuhan University. His current research mainly involves GNSS precise data processing and multi-sensor fusion.
	
	Contact him via	e-mail: \emph{xxli@sgg.whu.edu.cn}
\end{IEEEbiography}
\vspacebio
\begin{IEEEbiography}[{\includegraphics[width=1in,height=1.25in,clip,keepaspectratio,cframe={black!8!white}]{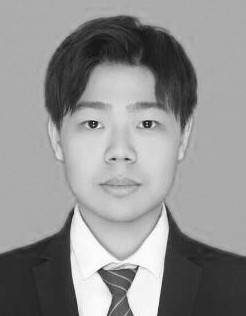}}]{Shengyu Li}
	received the M.S. degree in	geodesy and survey engineering from Wuhan University, Wuhan, China, in 2022.
	
	He is currently a doctor candidate at the School of Geodesy and Geomatics (SGG), Wuhan University, China. His current research focuses on multi-sensor fusion and integrated navigation system.
	
	Contact him via	e-mail: \emph{lishengyu@whu.edu.cn}
\end{IEEEbiography}
\vspacebio
\begin{IEEEbiography}[{\includegraphics[width=1in,height=1.25in,clip,keepaspectratio,cframe={black!8!white}]{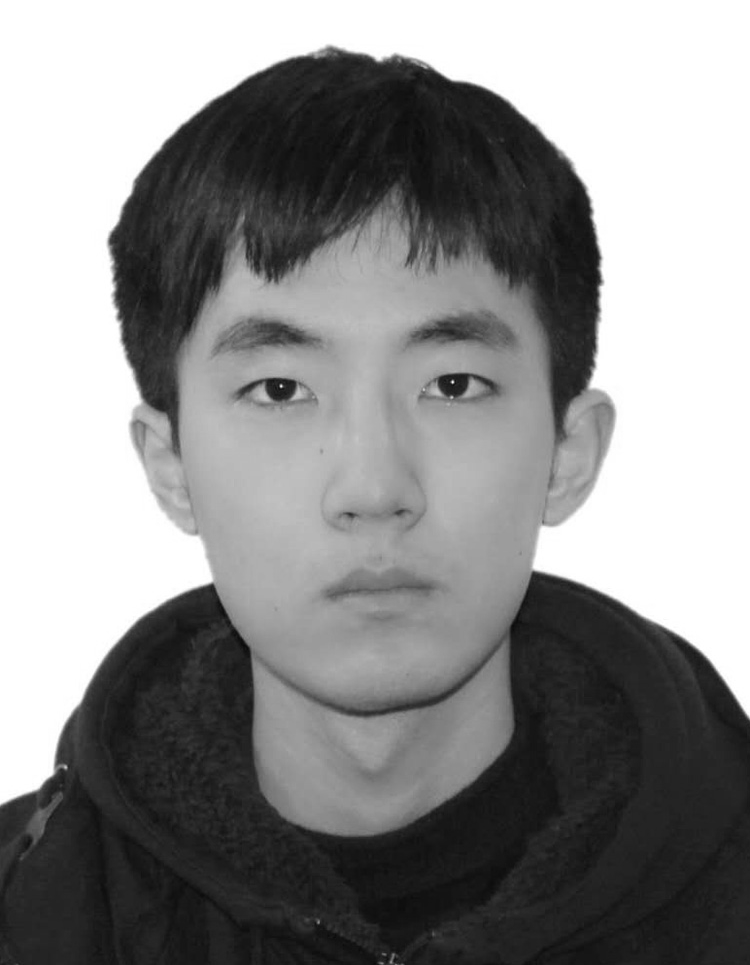}}]{Yuxuan Zhou}
	received the M.S. degree in geodesy and survey engineering from Wuhan University, Wuhan, China, in 2022.
	
	He is currently a doctor candidate at the School of Geodesy and Geomatics (SGG), Wuhan University, China. His current research areas include integrated navigation systems, simultaneous localization and	mapping, and multi-sensor fusion.
	
	Contact him via	e-mail: \emph{yuxuanzhou@whu.edu.cn}
\end{IEEEbiography}

\end{document}